\pdfoutput=1

\documentclass[11pt]{article}

\usepackage[final]{acl}

\usepackage{times}
\usepackage{latexsym}

\usepackage[T1]{fontenc}

\usepackage[utf8]{inputenc}

\usepackage{microtype}

\usepackage{inconsolata}

\usepackage{graphicx}

\usepackage{caption}
\usepackage[separate-uncertainty=true]{siunitx}
\usepackage{mathtools}
\usepackage{graphicx}
\usepackage{courier}
\usepackage{epstopdf}
\usepackage{color}
\usepackage{csquotes}

\usepackage{amsmath}
\usepackage{amsfonts}
\usepackage{amssymb}
\usepackage[subnum]{cases}
\usepackage{booktabs} 
\usepackage{pifont}

\usepackage{amssymb}
\usepackage{pifont}

\usepackage {tikz}
\usetikzlibrary {positioning}
\definecolor {processblue}{cmyk}{0.96,0,0,0}
\usepackage{lipsum,adjustbox}

\usepackage[subnum]{cases}
\usepackage{color,soul}

\usepackage{subcaption}

\usepackage{footnote}
\makesavenoteenv{tabular} 

\usepackage{multirow}
\usepackage{booktabs,subcaption,amsfonts,dcolumn}

\usepackage{xcolor}

\usepackage{textcomp}

\usepackage{bbold}
\usepackage{mathtools}
\usepackage{breqn}

\usepackage{amsthm}

\usepackage{makecell}

\usepackage[many]{tcolorbox}
\newtcolorbox{boxA}{
    fontupper = \bf,
    boxrule = 1.5pt,
    colframe = blue 
}

\title{Presentations are not always linear! GNN meets LLM for Document-to-Presentation Transformation with Attribution}

\author{Himanshu Maheshwari,$^{*}$ Sambaran Bandyopadhyay,$^{*}$ Aparna Garimella, Anandhavelu Natarajan \\
   Adobe Research \hspace{0.2cm} \\
  \texttt{\{himahesh, sambaranb, garimell, anandvn\}@adobe.com}\\
}

\begin{document}
\maketitle
\def\thefootnote{*}\footnotetext{First two authors contributed equally to this work}
\begin{abstract}
Automatically generating a presentation from the text of a long document is a challenging and useful problem. In contrast to a flat summary, a presentation needs to have a better and non-linear \textit{narrative}, i.e., the content of a slide can come from different and non-contiguous parts of the given document. However, it is difficult to incorporate such non-linear mapping of content to slides and ensure that the content is faithful to the document. LLMs are prone to hallucination and their performance degrades with the length of the input document. Towards this, we propose a novel graph based solution where we learn a graph from the input document and use a combination of graph neural network and LLM to generate a presentation with attribution of content for each slide. We conduct thorough experiments to show the merit of our approach compared to directly using LLMs for this task.
\end{abstract}

\section{Introduction}\label{sec:intro} 
Presentations are a very effective medium of communication in several day-to-day and business workflows.
Compared to a flat summarization, generating a presentation is more complex because it should have a nice narrative and coherence in the content along with the ability to convey the core ideas to the audience. This makes generating presentation a very tedious process for humans \cite{reynolds2011presentation}.
Document-to-slide generation using automatic methods has been garnering attention for several years.
These methods include using handcrafted and pre-defined heuristics or web schemas \cite{1565545,Winters2019AutomaticallyGE}. There are approaches which generate the agenda (\textit{i.e.,} sequence of slide titles) of a presentation based on the sections present in the document \citep{hu2013ppsgen,wang2017phrase} or require users to provide an agenda \citep{sun-etal-2021-d2s,li2021towards} and subsequently generate the slides as a single-document query-based summarization. However, manually coming up with an agenda is a difficult task, particularly for documents in the range of 10s of pages.

\begin{figure}[t!]
    \centering
    \includegraphics[width=\columnwidth]{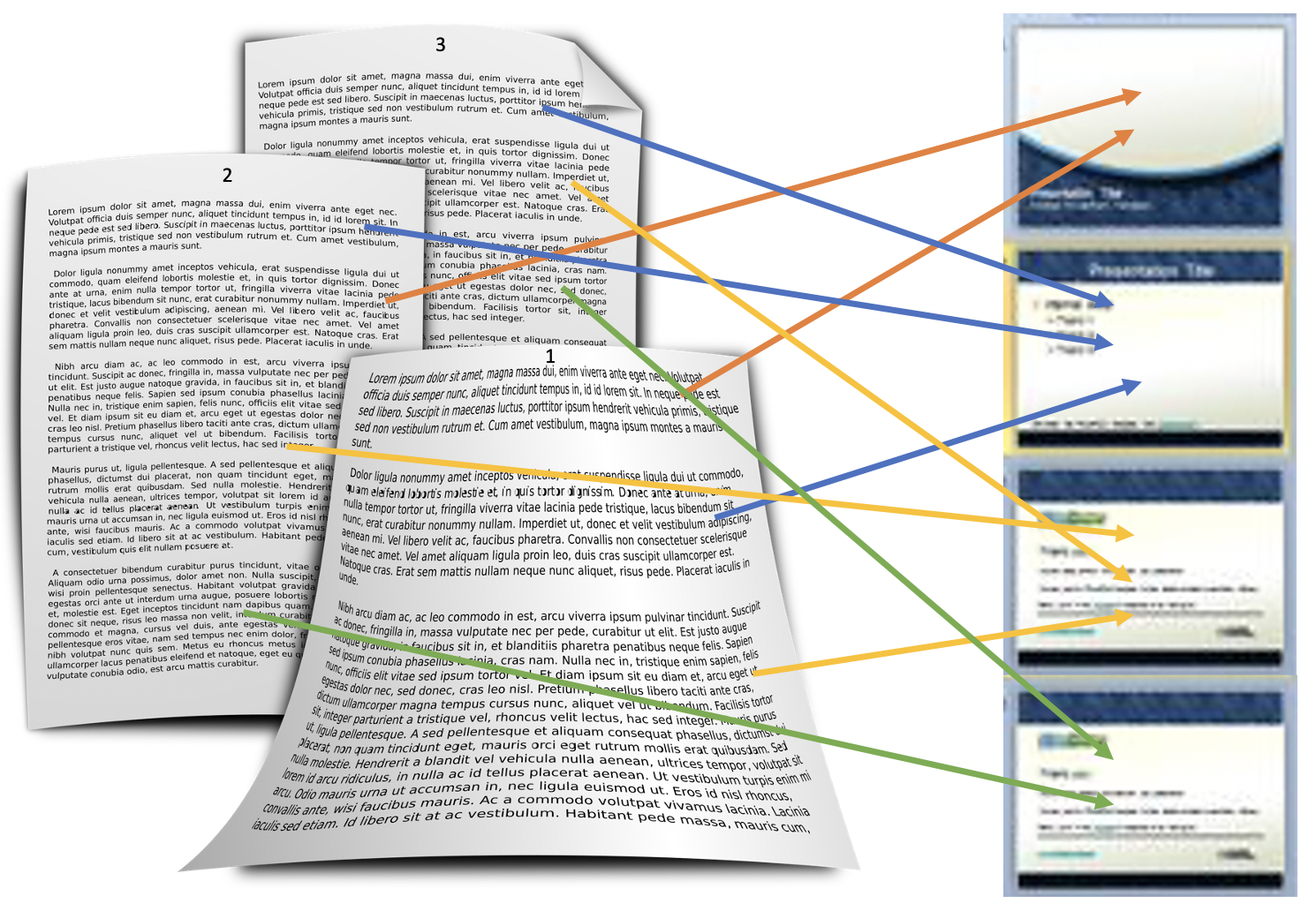}
    \caption{A presentation (right) with non-linear narrative and attribution to the source document (left)}
    \label{fig:non-linearity}
\end{figure}

This issue can be mitigated to an extent given the recent advances in generative language models \cite{NEURIPS2020_1457c0d6,touvron2023llama} with increasing context limits, in which we can prompt a large language model (LLM) to ingest the entire document context and generate an outline and the text content for a presentation. However this approach has three major limitations. First, the LLMs tend to hallucinate more and they often ignore the middle portion of a context as the context length grows \cite{liu2023lost}. This can be a serious issue if we want to generate slides from large documents and ensure good coverage of all the important concepts in the presentation. Second, processing the entire document in its reading order by an LLM often results in a summary-like overview of the document, as opposed to capturing a narrative-centric view that is required for a presentation.
Narratives (flow of information in the form of a story) \citep{xie-riedl-2024-creating} in presentations can be {\it non-linear} in nature, {\it i.e.,} paragraphs from across multiple sections of an input document contribute to a slide, and this is not necessarily in the linear reading order of the document (refer to Figure \ref{fig:non-linearity}). As shown in Section \ref{sec:non-linearity}, \textit{non-linearity} in presentations generated by the authors for the set of research papers in SciDuet dataset \citep{sun-etal-2021-d2s} is 38.6\%, whereas the ones generated by a GPT-based clustering baseline (GDP-GPT in Section \ref{sec:baseline}) has only 1.2\%. 
To generate these narratives, the non-linear relationship between the various pieces of content in the given document needs to be captured. Moreover, LLMs do not attribute the source content (\textit{e.g.}, a paragraph) for each part (\textit{e.g.}, a slide) of the generated content. This attribution is necessary to improve the reliability of the generated presentation and for further editing.

One can potentially think of posing the problem of non-linear way of generating presentation as a classification task of classifying a sequence of text elements (say, paragraphs) to one of the $K$ classes where each class represents a slide; or a clustering task of cluster the paragraphs to $K$ clusters. However, the number of slides needed from a document cannot be fixed over a set of documents. It can even vary for the same document depending on the audience, intent and the duration of presentation. Thus, it is not possible to pose it as a $K$-class classification task. Also, clustering as an unsupervised task can cluster text based on multiple aspects such as frequency of common words, common sub-topics, etc., where each generated cluster does not match to a slide.

\begin{figure*}[t!]
    \centering
    \includegraphics[width=\textwidth]{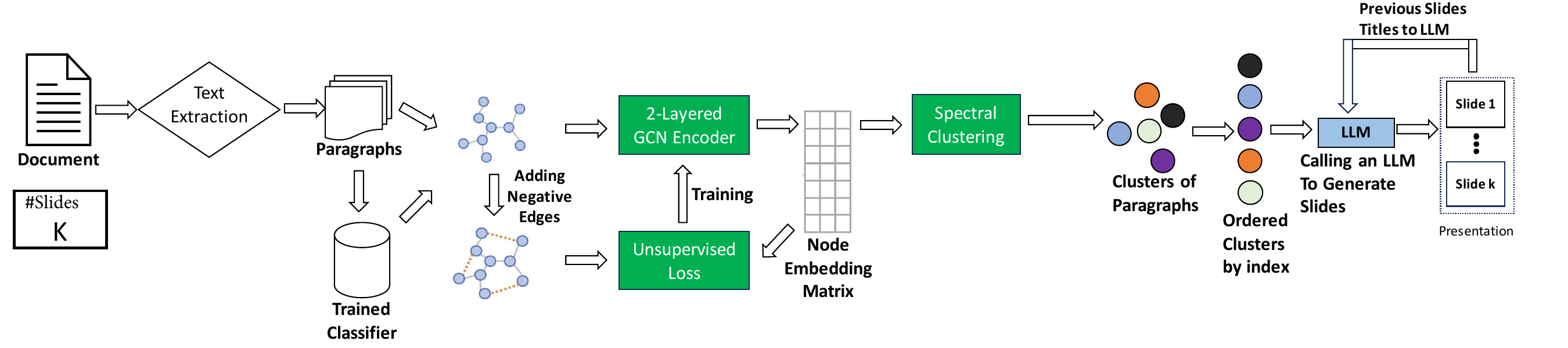}
    \caption{Architectural Diagram.}
    \label{fig:architecture}
\end{figure*}

To address the research gaps mentioned above, we propose a novel method of generating text presentation from a long input document as shown in Figure \ref{fig:architecture}. Our motivation is to infer the structure present between the text elements (\textit{i.e.,} paragraphs) of a document via corresponding latent slides (as shown in Figure \ref{fig:non-linearity}) by a learnable graph.
Following are the contributions made in this paper: \\
1. Automatically generating a non-linear presentation integrated with the content attribution from a given long document is a novel task to the best of our knowledge. \\
2. We propose a novel approach, referred as \textit{GDP} (\textit{G}raph based automated transformation of \textit{D}ocuments to \textit{P}resentation), which uses a combination of graph neural network (GNN) and LLM. Our method by design is able to capture the non-linearity in generating the presentation and it attributes the source paragraphs for each generated slide within the presentation. \\
3. We propose an evaluation framework which includes both automated and human evaluated metrics for document to presentation transformation. Our analysis shows the merit of GDP over the approaches that directly use SOTA LLM along with intelligent prompting techniques.

\section{Problem Formulation}\label{sec:prob} 
We are given with a training dataset which is a set of documents and the corresponding presentation slides. Let us denote this dataset as $\mathcal{D} = \{(D_1,P_1), (D_2,P_2), \cdots, (D_N, P_N)\}$, where $D_i$ is a document consisting of a sequence of $n_i$ text paragraphs as $D_i = (p_{i1}, \cdots, p_{in_i})$, $\forall i \in [N] =\{1,\cdots,N\}$. These paragraphs are indexed following the reading order in the document. Similarly, $P_i$ is a human-generated presentation from the document $D_i$. A presentation is a sequence of slides. So, $P_i = (s_{i1}, \cdots, s_{ik_i})$. Please note that different documents can have different number of paragraphs and the corresponding presentations can have different number of slides. As discussed in Section \ref{sec:intro}, we consider both the input document and the generated slides to contain only text. Ideally, the presentation should cover all the important aspects of the input document, with a nice flow of information such that it is easy to follow by a broader audience. Given this data, our goal is to generate a presentation for each document present in a test set $\mathcal{D'} = \{D_{N+1}, D_{N+2}, \cdots, D_{N+T}\}$. The number of slides $K_i$ to be generated from the test document $D_{i}$, $\forall i \in \{N+1,\cdots,N+T\} = [N \setminus T]$ is a user input during the inference time and we can not assumed this to be a constant over all the test documents. The training set and the test set of documents may come from the same distribution or from different distributions to test the generalizability of our proposed approach.

\section{Solution Approach}\label{sec:soln}
Following are the details of different components of our proposed approach GDP. Figure \ref{fig:architecture} shows the overall architecture.
%
We assume the input documents are in pdf format. We use a publicly available PDF Extract API~\footnote{\url{https://developer.adobe.com/document-services/apis/pdf-extract/}} to extract the text content from the documents. The output of extract is processed in such a way that we have the section and subsection titles and the text in the form of paragraphs within each section or subsection. The sequence of text elements is the same as the reading order of text in the pdf. We are not considering images, tables and other multimodal information present in the document in this work.

\subsection{Classifier: Dataset and Training} \label{sec:classifier}
The first step in our approach is training a classifier that can predict the probability of a pair of paragraphs from the given document going into the same slide. 
We use SciDuet dataset \citep{sun-etal-2021-d2s} which consists of document-presentation pairs, as discussed in section \ref{sec:dataset}. We leverage this dataset to create a synthetic dataset for training our classifier. 
We use sentence embedding \cite{reimers-2019-sentence-bert} for this. Let $\textbf{e}_{s_{ij}}$ be the embedding of $j$-th slide from $i$-th training presentation, and $\textbf{e}_{p_{ij}}$ be the embedding of $j^{th}$ paragraph from the corresponding document.
The set of selected paragraphs $P$ for a given slide $s_{ij}$ is determined as $\{p_{ij} \mid cosine(\textbf{e}_{s_{ij}}, \textbf{e}_{p_{ij}}) > \theta\}$.
We use a simple heuristic to define the threshold ($\theta$) for paragraph selection as $\theta = \{max\text{(C.S.)} - \frac{std\text{(C.S.)}}{2}\}$,
where $C.S.$ is a list of cosine similarity between $\textbf{e}_{s_{ij}}$ and the paragraphs in $D_i$, and $std$ is standard deviation of that. Paragraphs with a cosine similarity of less than $0.8$ are discarded to ensure a high-quality dataset. Additionally, we select a maximum of $10$ paragraphs per slide to ensure a balanced dataset. After this exercise, we get the lists of paragraphs that contribute together in the same slide within our training dataset.

\renewcommand{\arraystretch}{1.2}
\setlength{\tabcolsep}{4pt}
\begin{table}
\centering \scriptsize
\begin{tabular}{l|l|l|l}
\hline
\textbf{Dataset} & \textbf{\# Documents} & \textbf{\# Positive Samples} & \textbf{\# Negative Samples}\\
\hline
Training & 500 & 4,245 & 15,755 \\
Validation & 80 & 696 & 2,304 \\
Testing & 100 & 617 & 2,383 \\
\hline
\end{tabular}
\caption{Dataset Statistics for Classifier Training}
\label{tab:dataset_classifier}
\end{table}

Each pair of paragraphs in the list corresponding to a slide forms a positive sample for the classifier. For creating negative samples, for each paragraph, we sample ten random paragraphs from the document that never occurred with that paragraph. Note that by this approach, negative samples will be more than positive ones reflecting the real world scenario. After creating a dataset like this, we select $20,000$ samples for training and $3000$ for testing and validation. Please note that the documents used to create training, test, and validation datasets are mutually exclusive to prevent leakage. Please refer to Table \ref{tab:dataset_classifier} for the dataset details.
We fine-tuned a RoBERTa-base model \cite{liu2019roberta} on this dataset. Since the dataset is imbalanced, we use the standard weighted binary cross-entropy loss function. Best hyperparameters are found using a grid search on the validation set which gives an accuracy of $85.433\%$ and an F1 score of $0.778$ on the test set.

\subsection{Learning the Graph Structure}
\label{sec:learning_graph}
For the simplicity of notation, we will use $D = (p_1, p_2, \cdots, p_n)$ as the current document from which the presentation needs to be generated. The trained classifier from Section \ref{sec:classifier} will generate a pairwise probability $P(i, j)$ for any two paragraphs $p_i, p_j \in D$ which determines their chance of contributing to the same slide. 
Ideally, this probability is high when they actually contribute to the same slide and the probability is low when they do not contribute to the same slide. 
With this intuition, we aim to form a graph $G= (V, E, X)$ using these pairwise probabilities, where $V$ is the set of nodes, $E$ is the set of edges (undirected and unweighted) and $X$ is a node feature matrix.

As a simple heuristic, initially we create a node for each paragraph of the input document $D$. We connect two nodes $i$ and $j$ by an edge if the probability of them contributing to the same slide $P(i,j) > \alpha$, where $\alpha$ is a hyperparameter. It is well-understood that not all the paragraphs in a long document are covered in a presentation. To match this intuition and keep the graph small, we remove the isolated nodes from the graph. Each node (paragraph) is also associated with a vector (node feature) $x_i \in \mathbb{R}^F$ which is the corresponding paragraph embedding as discussed in Section \ref{sec:classifier}. Thus, the structure of the graph is heavily dependent of the trained classifier above. Next, we use a graph neural network to cluster the nodes of this graph so that each cluster can contribute to a slide.

\subsection{Slide Attribution via GNN and Clustering}
\label{sec:GNN}
Given the graph $G$ from the document $D$, we want to develop an unsupervised graph neural network which can obtain the vector representations (embeddings) of the nodes in such a way that when two nodes are directly connected in the graph, they have similar embeddings than two nodes which are far apart. Subsequently, a clustering algorithm is used on the generated node embeddings to find the clusters. Let us use $A \in \mathbb{R}^{|V| \times |V|}$ to denote the binary adjacency matrix of $G$, where $A_{ij} = 1$ if there is an edge between the nodes $i$ and $j$, otherwise $A_{ij}=0$. We use a 2-layered graph convolution encoder \citep{kipf2017semi} to obtain representation of each node as shown below:
\begin{align*} 
    Z = f(X, A) = \text{ReLU}(\hat{A} \; \text{ReLU}(\hat{A}XW^{(0)}) \; W^{(1)})
\end{align*}
where each row of $Z \in \mathbb{R}^{|V| \times F'}$ is the corresponding node embedding. We compute $\Tilde{A} = A + I$, where $I \in \mathbb{R}^{|V| \times |V|}$ is the identity matrix and the degree diagonal matrix $\Tilde{D_{ii}}$ with $\Tilde{D_{ii}} = \sum\limits_{j \in V} \Tilde{A}_{ij}$, $\forall i \in V$. We set $\hat{A} = \Tilde{D}^{-\frac{1}{2}} \Tilde{A} \Tilde{D}^{-\frac{1}{2}}$.
$W^{(0)}$ and $W^{(1)}$ are the trainable parameters of this GCN encoder.

Since the number of slides required for the same document can vary during the inference time depending on the need of a user, we cannot rely on any dataset to give a direct supervision to generate a presentation from a document. So,
We use an unsupervised loss function to train the parameters of the proposed GNN architecture. As our aim is to generate similar embeddings for the node pairs which are connected in the graph than any random pairs of nodes, we use the following strategy. For a given graph $G$, node embeddings are obtained by passing node feature matrix $X$ and the graph structure $A$ through the GCN encoder. Next, we randomly add a set of negative edges $E_n$, with $|E_n| = |E|$ in the graph. We pose the training task as to minimize the following binary cross entropy loss on positive and negative edges of the graph as shown below:
\begin{align*}\label{eq:dsf}
    \mathcal{L} =& - \sum_{(i,j) \in E} \log \Big(\sigma(z_i^Tz_j) \Big) \\ &- \sum_{(i,j) \in E_n} \log \Big( 1 - \sigma(z_i^Tz_j) \Big)
\end{align*}
Here $\sigma$ is the sigmoid function and $z_i$ is the $i$-th row of $Z$, i.e., the node embedding of the $i$ node in the graph. We use standard ADAM optimization technique \citep{kingma2014adam} with a learning rate $0.01$ to minimize the loss function above.

Once the unsupervised training is complete, we obtain the node embedding matrix $Z$ from the graph where each row is a paragraph embedding of dimension $F'$ from the given document. Since our main goal is to cluster the paragraphs in such a way that each cluster can correspond to a slide, we use spectral clustering \citep{ng2001spectral} on these paragraph embeddings from GNN. The number of clusters is kept as the number of slides $K$ required for the document $D$, and this number varies over the documents during inference. We have observed empirically that spectral clustering is able produce more balanced clusters compared to other algorithms such as KMeans on these node embeddings. At the end of this step, we obtain a clustering of paragraphs (nodes) of the current document as $\mathcal{C} = \{C_1,C_2,\cdots,C_K\}$. 

\subsection{Generating the Presentation}\label{sec:gen_presen}
Please note that the clusters obtained above are unordered. To be able to generate slides from the clusters, we first order them using a simple heuristic. For any cluster $C_k$, where $k \in [K]$, we consider all the paragraphs that belong to that cluster and take the minimum of their indices (please note that paragraph indices follow the reading order in the document as mentioned in Section \ref{sec:prob}). Mathematically, $index_C = \min\{i \vert p_i \in C\}$. We use this $index$ number to sort the clusters in increasing order and then associate the first cluster (with the smallest $index$ number) as the one corresponding to the first slide, and so on. Since, paragraph indices follow the reading order of the document, we wanted to roughly follow that in the generated presentation. One can see that non-linearity is there in mapping the paragraphs to slides since paragraphs from any part of the document can contribute to any slide of the presentation. Let us reorder the clusters and denote the clustering as $\mathcal{C} = (C'_1, C'_2, \cdots, C'_K)$, where they are sorted according to the $index$ discussed above.

Now, let us discuss the generation of the presentation $P = (s_1,s_2, \cdots, s_K)$ for the given document $D$. For a slide $s_k$, we know the corresponding cluster $C'_k$, and the paragraphs forming that cluster. We use GPT-3.5~\footnote{Please note that GDP can support any LLMs even with lesser context length since we feed only a few paragraphs at a time. The choice of GPT-3.5 was to support some of the baselines in Section \ref{sec:baseline} which need to see the entire document within a single prompt.} to generate the slides in sequence. To generate a slide $s_k$, we provide the texts present in the paragraphs $(p_i \vert p_i \in C'_k)$, along with the titles of the previous slides $s_1,\cdots,s_{k-1}$. Experimentally, we found that providing information about the previous slides help GPT to maintain a good flow in the presentation. The prompt for the generation of $k$th slide of the presentation is shown in Appendix \ref{sec:prompt_final_slide}. To generate the whole presentation $P$, we made $K$ such calls in sequence where $K$ is the required number of slides.



\section{Experiments}\label{sec:exp}

\subsection{Datasets}\label{sec:dataset}
We use the SciDuet dataset proposed in \citet{sun-etal-2021-d2s}, which has research papers and their presentations. The papers are from ICML, NeurIPS, and ACL. We use papers from ICML and NeurIPS as their PDFs are available. We split 500 papers for training, 80 for validation, and 100 for testing. This dataset is used to train the classifier (\S\ref{sec:classifier}), find the right hyperparameters, and perform final testing.


\subsection{Baseline Algorithms and Model Ablation}
\label{sec:baseline}
We use D2S model \citep{sun-etal-2021-d2s} and different ways of prompting GPT as baseline algorithms in our experiments. The exact prompts for all the baseline models are presented in the Appendix.  All baseline experiments use GPT3.5-turbo-1106 with a 16K context length.\\
1. \textbf{D2S}: We use D2S \citep{sun-etal-2021-d2s} as a semi-automatic baseline for this task. D2S takes the ground truth slide titles from the test set and extract content to generate the presentation. The authors did not provide checkpoints, training scripts, or datasets for their BERT-based IR model \footnote{\url{https://github.com/IBM/document2slides/issues/3}}. To encode queries, we opted for a pre-trained BERT model due to this absence. \\
2. \textbf{GPT-Flat}: 
In this baseline we use a simple prompt and provide the full input text from the document to GPT model to generate the presentation.\\
3. \textbf{GPT-COT}: \citet{wei2022chain} demonstrates that chain-of-thought (COT) prompting enhances LLM performance across tasks. Thus, in this baseline, we use a COT prompt for GPT-3.5 to generate presentations from documents.\\
4. \textbf{GPT-Constrained}: To optimize text per slide, we instruct GPT with a modified COT prompt, specifying the number of bullet points and the maximum words per bullet point for each slide in this baseline.\\
5. \textbf{GDP-KMeans}: This is a \textit{model ablation} for our proposed approach. This study replaces the proposed graph learning followed by clustering part with KMeans on paragraph embeddings from sentence transformers (\textit{all-mpnet-base-v2}) \cite{reimers-2019-sentence-bert}. The rest of the pipeline is the same as our approach. This is to study the impact of graph learning in our proposed pipeline.\\ 
6. \textbf{GDP-Agglo}: This is same as the \textit{model ablation} study above, but we use agglomerative clustering instead of KMeans.\\
7. \textbf{GDP-GPT}: This is another \textit{model ablation} of GDP. In GDP, the core process is classifying paragraph pairs, constructing a graph, and performing node clustering. In GDP-GPT, we replace this process with GPT-3.5. GPT-3.5 takes the text of the paragraphs and decides which paragraphs contribute to each slide. Thus, the mapping between paragraphs and slides is done directly by GPT-3.5. After that, we use the same prompts present in Section \ref{sec:gen_presen} to generate the final presentation.

\subsection{Experimental Setup}
We use a RoBERTa base model for a classifier with batch size $12$, a learning rate of $10^{-5}$, and a dropout of $0.4$. RoBERTa was finetuned on Nvidia A10 G for 3 hours. We use the \textit{all-mpnet-base-v2} sentence transformer model for all our experiments. We use the \textit{gpt-3.5-turbo-1106} model for all our experiments with a temperature of $0.7$ and \textit{top_p} of $0.95$. For G-Eval-based evaluation, we use the \textit{gpt-4} model with a temperature of $0.7$, \textit{top_p} of $0.95$, and the number of generations as $10$.

\renewcommand{\arraystretch}{1.4}
\setlength{\tabcolsep}{5pt}
\begin{table*}[ht]
\centering
\resizebox{\linewidth}{!}{
\begin{tabular}{c|ccc|cc|c|c}
    \toprule
      Method & \multicolumn{3}{c}{ROUGE-1 $\uparrow$} & \multicolumn{2}{c}{Coverage $\uparrow$}& PPL $\downarrow$ & G-Eval $\uparrow$ \\
      & Recall & Precision & F1 & Paragraph & Sentence & &\\
    \hline
    D2S & 29.533 $\pm$ 9.407 & 9.086 $\pm$ 3.764 & 12.97 $\pm$ 3.839 & 38.489 $\pm$ 5.432 & 24.247 $\pm$ 3.385 & 77.386 $\pm$ 28.958 & 5.134 $\pm$ 0.843\\
    GPT-Flat & 37.277 $\pm$ 16.779 & 41.151 $\pm$ 14.001 &34.620 $\pm$ 9.347 & 33.411 $\pm$ 8.123 & 22.833 $\pm$ 4.037 & 133.516 $\pm$ 96.926 & 7.974 $\pm$ 0.506\\
    GPT-COT & 38.010 $\pm$ 16.405 &	40.911 $\pm$ 12.529 & \textbf{35.504 $\pm$ 8.824}  & 34.830 $\pm$ 6.069 & 23.383 $\pm$ 4.076 & 104.140 $\pm$ 53.702 & \textbf{8.054 $\pm$ 0.584} \\
    GPT-Cons & \textbf{43.753 $\pm$ 18.118} & 33.652 $\pm$ 13.275 & 33.273 $\pm$ 9.683 & 34.598 $\pm$ 7.638	& 23.313 $\pm$ 4.171 & 121.372 $\pm$ 112.161 & 7.824 $\pm$ 0.709\\
    \hline
    GDP-KMeans & 22.92 $\pm$ 12.784 & \textbf{62.248 $\pm$ 12.723} & 30.71 $\pm$ 10.38 & 34.641 $\pm$ 5.198 & 22.168 $\pm$ 3.27 & 58.503 $\pm$ 12.762 & 7.204 $\pm$ 0.603\\
    GDP-Agglo & 22.933 $\pm$ 13.011 & 56.8 $\pm$ 12.702 & 29.611 $\pm$ 9.705 & 39.043 $\pm$ 5.454 & 23.879 $\pm$ 3.688 & 62.801 $\pm$ 19.144 & 7.411 $\pm$ 0.684\\
    GDP-GPT & 33.206 $\pm$ 19.354 & 48.392 $\pm$ 19.491 & 31.994 $\pm$ 10.933 & 38.262 $\pm$ 6.162 & 23.964 $\pm$ 3.696 & 57.230 $\pm$ 18.826 & 7.691 $\pm$ 0.529 \\
    \textbf{GDP} & 23.788 $\pm$ 14.103 & 58.583 $\pm$ 12.644 & 30.589 $\pm$ 10.353 & \textbf{39.050 $\pm$ 5.474} & \textbf{24.263 $\pm$ 3.230} & \textbf{56.005 $\pm$ 15.879} & 7.781 $\pm$ 0.439  \\
    \bottomrule
\end{tabular}
}
\caption{Performance analysis of the generated presentation with different automated metrics on test set.}
\label{tab:automated_results}
\end{table*}

\subsection{Automatic Evaluation Metrics}\label{sec:autom_metrics}
Since the task of document to presentation is under-explored in the literature, there is no established evaluation framework. We adopt the following metrics from ML and NLP literature, tailored to suit this task:\\
1. \textbf{ROUGE-1}: We use ROUGE-1\footnote{https://pypi.org/project/rouge-score/} to compare the text of the generated presentations with the ground truth presentation. We choose ROUGE-1 because it looks at individual words. Since GPT often produces long responses, we check at the word level to see if the key terms are there in the final result.\\
2. \textbf{Coverage}: This is an unsupervised metric that intuitively captures the ``coverage'' of the content of a super set in a subset \citep{kothawade2020deep}. For this task, we define Coverage (paragraph / sentence levels) to be the average cosine similarity of a slide (bullet point) from the presentation and a paragraph (sentence) from the document based on their sentence embeddings \cite{reimers-2019-sentence-bert}. Clearly, more is the Coverage, better is the quality (more details in the Appendix).
%
\\
3. \textbf{Perplexity (PPL)}: Perplexity is a metric to indicate the fluency of the generated text. It is obtained using GPT-2, as discussed in \citet{liu-etal-2021-learning}. Perplexity measures how likely the language model (GPT-2 here) is to generate the sequence. If GPT-2 assigns a high probability to the token present in the sentence, the perplexity will be lower, indicating a fluent and grammatically correct sentence.\\
4. \textbf{G-Eval for presentation quality}: G-Eval \cite{liu-etal-2023-g} is a well-established metric that uses GPT-4 to evaluate various NLP tasks. It has a very high correlation with humans. We use G-Eval to measure the overall presentation quality in terms of organization, effectiveness, clarity, coherence, and the ability to convey complex ideas to the audience. Please refer to the appendix for the exact prompt.

\subsection{Results and Analysis}
We have presented the results of the baseline algorithms and GDP with its model variant on the test set (discussed in \S\ref{sec:dataset}) in Table \ref{tab:automated_results}. We can make the following observations: \textbf{(1)} We can see that recall of GDP for ROUGE-1 is significantly less compared to GPT based baselines. 
However, GDP is able to achieve very high ROUGE-1 Precision, beating all baselines but GDP-KMeans. As also discussed in \citet{sun-etal-2021-d2s}, ROUGE is not the best metric to evaluate presentations as multiple correct presentations differ at the lexical level, thus having different ROUGE1 scores. Our results built trust that GDP outputs important words in the final presentation. \textbf{(2)} GDP performs the best in terms of both paragraph-level and sentence-level coverage. This shows that GDP covers the entire document and does not miss out on some sections, a problem that human annotators also identified with the baselines. \textbf{(3)} GDP and its variant GDP-KMeans and GDP-GPT achieve a very low score of PPL (which indicates better performance) compared to baselines. This means clustering the paragraphs, generating a slide from each cluster, and using suitable prompts ensures a smooth flow of text and information in the presentations generated by GDP.  \textbf{(4)} Finally for G-Eval, performance of all the GPT based algorithms, GDP-GPT and GDP are very close. GDP-KMeans perform poorly on G-Eval, providing trust in our algorithm of clustering paragraphs. However, the performance of D2S is not good for most of the metrics showing the importance of LLMs for generating content.


Appendix \ref{sec:qualitative} shows some qualitative analysis of a presentation generated by our proposed approach and compare that with the one generated by a baseline from the same input document.

\subsection{Evaluation of Non-linearity}\label{sec:non-linearity}
Section \ref{sec:intro} motivates the need of non-linearity in the generated presentation. In this section, we propose a metric to quantify the amount of non-linearity present in ground truth document and the ones generated by our proposed algorithm.
Suppose the given document is $D = (p_1,p_2,..,p_n)$ where $p_i$ is a paragraph within it and the corresponding presentation is $P = (s_1,s_2,…,s_k)$, where $s_j$ is a slide. 
As discussed in Section \ref{sec:prob}, the paragraphs and slides are arranged as in the reading order for the document and the presentation respectively. We also assume that we know the source paragraphs from the document for each slide in the presentation through some attribution mechanism. For example, consider the source paragraphs for a specific document and the corresponding presentation as: $s_1 \xleftarrow{} (p_1, p_3)$; 
$s_2 \xleftarrow{} (p_2, p_5, p_7)$ and so on. 
If we consider each slide in order and put the indices of the corresponding source paragraphs in order, we will get a sequence $S$ of integer numbers. For the above example, the sequence will be: $S = (1, 3, 2, 5, 7)$. If the generated presentation is completely linear (in terms of information flow) with respect to the document, the sequence will exactly follow the natural ordering of the integers. So, if the number $i$ has occurred before the number $j$ in this sequence, $i < j$ (assuming there is no repetition of paragraphs across multiple slides). However, this may not be true when the presentation is non-linear. So \textbf{Non-linearity} present in the generated presentation is proportional to the number of pairs of numbers in the sequence for which the natural ordering is not maintained. Mathematically, 
    $\text{Non-linearity} = \sum_{\substack{i,j \in S \\ ind(i)<ind(j)}} \frac{\mathbb{1}(i>j)}{{|S| \choose 2}} \times 100 \%$,
where $\mathbb{1}$ is the indicator function. More is the value of the metric Non-linearity, more is the non-linearity in the generated slide. For example, value of Non-linearity for the sequence $S=(1, 3, 2, 5, 7)$ is $1/5C2 \times 100\% = 10\%$. 

We compute and find that the value of non-linearity of the ground truth presentations present in the SciDuet dataset which includes research papers from ICML, NeurIPS, and ACL is $38.6\%$ (based on the attribution mechanism to generate the pseudo ground truth to link a slide with a set of paragraphs in Section \ref{sec:classifier}). Since, GDP, GDP-KMeans, and GDP-GPT generate attribution as a bi-product of generating slides, we compute the non-linearity of the generated slides by these approaches. It turns out that the Non-linearity of GDP is $24.9\%$, Non-linearity of GDP-KMeans is $39.7\%$, and Non-linearity of GDP-GPT is $1.2\%$. We could not compute the values for other baselines since attribution is not a direct output from those algorithms.

Following are the observations from this study:
(1) Human generated presentations are highly non-linear ($38.6\%$) in nature.
(2) GDP-KMeans had a very high non-linearity of $39.7\%$, even higher than human-generated presentations. On manual inspection, we found that it is clustering some very random paragraphs together, which is undesirable.
(3) Since, GDP-GPT uses GPT-3.5 to cluster the paragraphs and it is known that GPT tends to follow the ordering of the text present in the context \citep{liu2023lost}, GPT based approaches are inherently quite linear in nature (with a Non-linearity of $1.2\%$ for GDP-GPT).
(3) The construction of graph using the results of the classifier and the subsequent use of GNN and clustering makes GDP quite non-linear ($24.9\%$) in nature. Graphs are indeed very good to handle non-linearity. Thus, the presentations generated by GDP is more close to human made presentations.

\begin{table*}[ht]
\centering
\resizebox{\linewidth}{!}{
\begin{tabular}{c|cccccccc}
    \toprule
      Method & Language & Slide Uniformity & Coverage & Non-repetition & Narrative & Consistency & Attribution & Utility \\
    \midrule
    GPT-Flat & 2.300 $\pm$ 1.160 & 2.200 $\pm$ 1.135 & 1.300 $\pm$ 0.483 & 2.800 $\pm$ 1.549 & 1.700 $\pm$ 0.675 & 4.400 $\pm$ 0.5116 & NA & 1.200 $\pm$ 0.632 \\
    GPT-COT & 2.300 $\pm$ 1.160 & 2.400 $\pm$ 1.350 & 1.500 $\pm$ 0.850 & 2.800 $\pm$ 1.549 & 1.700 $\pm$ 0.675 & 4.400 $\pm$ 0.5116 & NA & 1.200 $\pm$ 0.632 \\
    GPT-Cons & 2.300 $\pm$ 1.160 & 2.000 $\pm$ 1.054 & 1.100 $\pm$ 0.316 & 2.800 $\pm$ 1.549 & 1.700 $\pm$ 0.675 & 4.400 $\pm$ 0.5116 & NA & 1.200 $\pm$ 0.632 \\
    \midrule
    GDP-GPT & 3.300 $\pm$ 0.180 & 2.600 $\pm$ 0.430 & 3.200 $\pm$ 0.200 & 3.300 $\pm$ 1.100 & 3.000 $\pm$ 0.500 & \textbf{4.750 $\pm$ 0.463} & 2.300 $\pm$ 1.300 & 3.425 $\pm$ 0.446 \\
    \textbf{GDP} & \textbf{4.125 $\pm$ 0.991} & \textbf{3.875 $\pm$ 0.835} & \textbf{4.000 $\pm$ 0.926} & \textbf{4.000 $\pm$ 0.000} & \textbf{3.750 $\pm$ 1.165} & \textbf{4.750 $\pm$ 0.463} & \textbf{2.500 $\pm$ 1.915} & \textbf{4.625 $\pm$ 0.744}  \\
    \bottomrule
\end{tabular}
}
\caption{Human evaluation on the generated presentations on five \textbf{research papers}}
\label{tab:human_rsults_research}
\end{table*}

\begin{table*}[ht]
\centering
\resizebox{\linewidth}{!}{
\begin{tabular}{c|cccccccc}
    \toprule
      Method & Language & Slide Uniformity & Coverage & Non-repetition & Narrative & Consistency & Attribution & Utility \\
    \midrule
    GPT-Flat & 3.143 $\pm$ 0.949 & 4.500 $\pm$ 1.286 & 3.286 $\pm$ 1.541 & 4.643 $\pm$ 1.406 & 4.143 $\pm$ 1.406 & 4.929 $\pm$ 0.267 & NA & 3.857 $\pm$ 1.351 \\
    GPT-COT & 3.786 $\pm$ 0.802 & \textbf{5.000 $\pm$ 0.000} & 4.143 $\pm$ 1.292 & 4.857 $\pm$ 0.363 & 4.929 $\pm$ 0.267 & 4.929 $\pm$ 0.267 & NA & 4.571 $\pm$ 0.756 \\
    GPT-Cons & 3.643 $\pm$ 0.929 & 4.500 $\pm$ 1.286 & 3.500 $\pm$ 1.743 & 4.571 $\pm$ 1.089 & 4.357 $\pm$ 1.336 & 4.929 $\pm$ 0.267 & NA & 4.071 $\pm$ 1.328 \\
    \midrule
    GDP-GPT & 4.643 $\pm$ 0.842 & 4.714 $\pm$ 0.726 & 4.429 $\pm$ 1.453 & 4.500 $\pm$ 0.760 & 4.786 $\pm$ 0.426 & \textbf{5.000 $\pm$ 0.000} & \textbf{5.000 $\pm$ 0.000} & 4.786 $\pm$ 0.579 \\
    \textbf{GDP} & \textbf{4.857 $\pm$ 0.363} & 4.929 $\pm$ 0.267 & \textbf{4.714 $\pm$ 0.825} & \textbf{4.714 $\pm$ 0.611} & \textbf{5.000 $\pm$ 0.000} & \textbf{5.000 $\pm$ 0.000} & \textbf{5.000 $\pm$ 0.000} & \textbf{4.929 $\pm$ 0.267} \\
    \bottomrule
\end{tabular}
}
\caption{Human evaluation on the generated presentations on seven \textbf{business documents}}
\label{tab:human_rsults_business}
\end{table*}

\subsection{Human Evaluation}\label{sec:human_eval}

Presentation quality is subjective, and there is no universally defined \textit{best presentation} for a given document \citep{sun-etal-2021-d2s}. We conduct a comprehensive human evaluation to further understand the presentations generated by our approach and by some selected baselines based on their performance on manual inspection and the available budget. 

For this task, we discussed with subject matter experts and selected two sets of documents: \textbf{(1)} Five \textbf{research papers} 
\textbf{(2)} Seven \textbf{business documents} comprising of technical manuals, reports, and news articles. This is a domain shift from the training set.
Following are the metrics we have used for human evaluation:
1. Quality of \textbf{Language}; 2. \textbf{Slide Uniformity} to check the alignment between the title and main text, as well as coherence and uniformity within the slide's content; 3. 3. \textbf{Coverage} of the content; 4. \textbf{Non-repetition} of content across the slides; 5. Quality of \textbf{Narrative} (the flow of information) in the presentation; 6. \textbf{Consistency} by not including any content (or / and hallucination) outside of the given document; 7. \textbf{Attribution Quality}; and 8. \textbf{Utility} to check how easily a user can use / update a generated presentation. More details can be found in the Appendix. 
We use a Likert scale from 1 to 5 for all the metrics.

We hired professional human reviewers~\footnote{https://www.upwork.com/}, two with research backgrounds for evaluating presentations on research papers and two with experience in professional writing for business documents. We explained the metrics and the evaluation process to them over multiple sessions. They had no knowledge of the algorithms used to prevent preconceived bias. 
Each reviewer rated each presentation on a scale of 1 to 5 for each metric while also providing explanations. The cohen kappa score for inter annotator agreement is 0.386.


\paragraph{Analysis of the Results} Anonymous link to the selected documents for human evaluation and the corresponding generated presentation can be found in the Appendix. We show the average and standard deviation of ratings by 2 reviewers for five \textbf{research papers} in Table \ref{tab:human_rsults_research}. GDP significantly outperforms baselines and its variant across all metrics. For the GPT based algorithms, a few common concerns were \textit{``title matches the text in the slide, but the slides go off topic often''}, \textit{``There are several details that have not been touched at all, like data collection, annotation, unigram, part-of-speech tag..''} and \textit{``there was no significant narration that was conveyed in the ppt''}. 
This aligns with our intuition from Section \ref{sec:intro} that GPT struggles with lengthy input contexts. Particularly for research papers, where discussions focus on a single topic with repeated words and concepts, GPT-based algorithms struggle to produce quality output despite various prompting techniques. Whereas for GDP, reviewers appreciate the coverage (\textit{``The reason is simply because all the data was covered by the slides''}), non-repetition (\textit{``data provided in the slides wasn’t repeated}), consistency and attribution (\textit{``no hallucination ''}). There was also some concern on GDP about the depth of the generated presentation (\textit{``The deck covers a lot of content but doesn't deep dive''}).

The human evaluation results for \textbf{business documents} are presented in Table \ref{tab:human_rsults_business}. The results highlight GDP's ability to generalize to a new domain, outperforming all algorithms in all metrics except slide uniformity. Unlike research papers, business documents have shorter text on average. All the algorithms perform good on building narratives and maintaining information consistency in presentations.  However, the reviewers are not satisfied with the language, coverage and utility of the presentations generated by the GPT based algorithms (\textit{``The presentation is not an ideal first draft as it very briefly summarizes the content of the input document with limited accuracy and consistency''}). But they do appreciate GDP for these metrics (\textit{``The presentation is an efficient first draft ''}).

\section{Discussions and Conclusion}
This paper presents an end-to-end novel approach, GDP, for transforming a long document into a text presentations. GDP employs a classifier to build a document graph, followed by graph neural networks and clustering. It then uses an LLM to generate slides from each paragraph cluster. We propose evaluation frameworks, and the results indicate several drawbacks of directly using GPT-based approaches with different prompting techniques. The evaluation shows that GDP can automatically generate a presentation that serves good as a first draft.

\section{Limitations}
In this work, we focus only on the text part of an input document and generate only text presentation. Multimodal content such as images, diagrams and tables carry important information present in a document. Handling such multimodal content and also extract or generate more of them in the output presentation is an interesting problem. We seek to use a vision language model such as CLIP \citep{radford2021learning} or a large multimodal model such as LLaVA \citep{liu2024visual} in our pipeline for this task in some future work.

Another important aspect of a presentation is the selection of a relevant template and layout that goes well with the content. For example, a presentation with a formal content should have a different background and colours than the one with a very casual content. Currently in our implementation, we use a default vanilla template for all the generated presentations. Selection or recommendation of a suitable template and layout for a presentation is out of scope for this work and can be addressed in future.


\bibliography{GDP}

\newpage
\appendix

\section*{Appendix}

\section{Qualitative Analysis with Generated Examples}\label{sec:qualitative}
\begin{figure*}[t!]
    \centering
    \includegraphics[width=\textwidth]{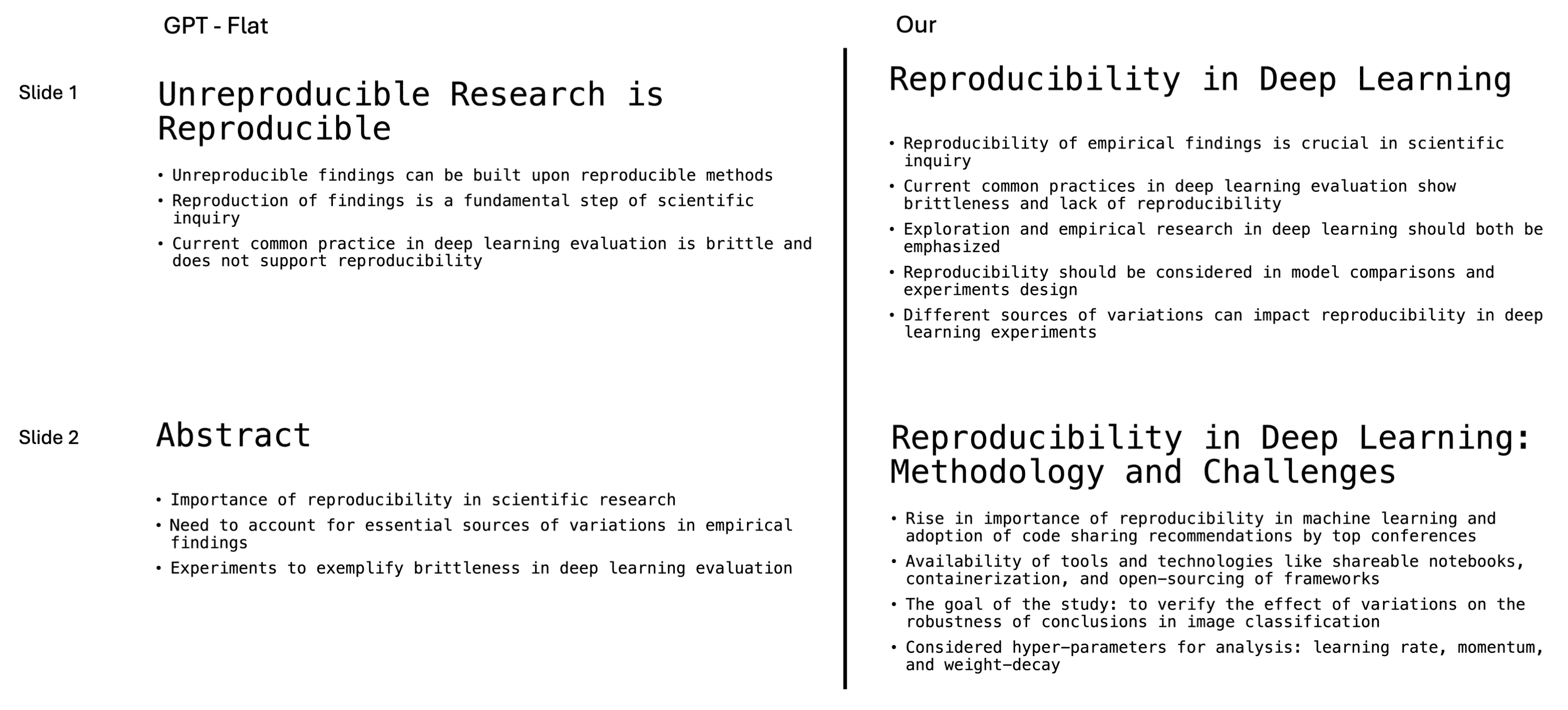}
    \caption{Qualitative example to compare the slides generated by a baseline GPT-Flat and our approach GDP from the input document \url{https://proceedings.mlr.press/v97/bouthillier19a/bouthillier19a.pdf}.}
    \label{fig:qualitative_example}
\end{figure*}

Figure \ref{fig:qualitative_example} displays the first two slides generated by GPT-Flat and Our approach, respectively. The source document is a randomly selected paper from the test set (available at \href{https://proceedings.mlr.press/v97/bouthillier19a/bouthillier19a.pdf}{this link}). Our approach creates a presentation with a smoother transition between slides, both in terms of heading and content, thus having better flow. These slides are more engaging compared to those generated by GPT-Flat, which tend to resemble the structure of a paper and have headings such as `Abstract'. The second slide from our approach is non-linear in nature. It combines relevant content from paragraph 3 in Introduction, section 3, and paragraphs 3-5 from section 7 of the input paper. This enables the generated slide to summarize the key points from different parts of the paper.

\section{Related Work}\label{sec:rw} 
\textbf{Document transformation}:
Assuming the slide titles to be the same as the document sections, there are works which use a query specific summarizer \citet{sravanthi2009slidesgen}, learn sentence importance \citep{hu2013ppsgen} and extract hierarchical relations between phrases \citep{wang2017phrase} to generate the presentation.
\citet{sun-etal-2021-d2s} addressed the document-to-slides generation task as a query-based single-document text summarization.
With the same motivation, \citet{li2021towards} learns two sentence extractors collaboratively and bootstrap the performance of each other.
All these approaches still require the user to come up with the agenda of slides. \citet{Fu2021DOC2PPTAP} proposed a hierarchical sequence-to-sequence model with trainable policy to summarize a given document into a structured slide deck. However, their approach needs larger training data and is domain specific.

\textbf{Document summarization}: 
A presentation is often viewed as a summary of the given document. Initial work \cite{7877699,7975205,Wang_Wan_Du_2017} used extractive summarization to create slides. 
Another related area, Multimodal, and abstractive summarization works such as \cite{zhu-etal-2018-msmo,li-etal-2020-vmsmo} proposed joint modeling of text and image inputs to generate multimodal summaries in an abstractive manner. However, as discussed in Section \ref{sec:intro}, a summary-based viewpoint for generating presentations fails to take into account the high-level ``narrative" that is integral to a document like presentation.


\textbf{LLMs for presentation generation}:
There has been a surge in various AI-based startups to generate presentations given a simple prompt leveraging the generative power of LLMs. However, creating a slide deck for an existing document using LLMs has been under-explored.

\section{Automated Metric: Coverage}
It is an unsupervised metric which intuitively capture how much does a subset ``cover'' the content of the super set \citep{kothawade2020deep}. In literature, it has been used for extractive summarization . We use the following definition of Coverage (at \textbf{paragraph} to slide level) in this work:
\[ Coverage = \frac{\sum_{\mathbf{e}_p \in D}\sum_{\mathbf{e}_s \in P}cosine\bigl(\mathbf{e}_p, \mathbf{e}_s \bigr)}{|D||P|} \times 100 \%\]
Here, $\mathbf{e}_p$ is a paragraph embedding from the given document and $\mathbf{e}_s$ is a slide embedding from the generated presentation as obtained by a sentence transformer model \cite{reimers-2019-sentence-bert}.
If some paragraph is similar to multiple slides, it is covered well in the overall presentation. 
Similarly, coverage can also be computed ta \textbf{sentence} level by replacing a paragraph with a sentence from the  document and a slide with a bullet point (or sentence) from the presentation in the equation above. Sentence level coverage offers a finer granularity than paragraph-level coverage. Higher coverage indicates a better presentation.

\section{Human Evaluation Metrics}
Following are the details of the metrics used in human evaluation conducted in this paper.
\\ 
1. \textbf{Language}: This evaluates slide text quality based on language correctness, text quantity, and bullet point length.\\
2. \textbf{Slide Uniformity}: This metric evaluates slide coherence, ensuring alignment between the title and main text, as well as coherence and uniformity within the slide's content.\\
3. \textbf{Coverage}: This metric captures if all the important concepts in the given document are captured nicely in the generated presentation.\\
4. \textbf{Non-repetition}: This metric makes sure the presentation doesn't repeat things and has different information on each slide.\\
5. \textbf{Narrative}: This metric evaluates how well the presentation tells a story by checking the smooth flow of information. It looks for a well-organized order of slides to ensure a cohesive story.\\
6. \textbf{Consistency}: LLMs are prone to hallucination by sometimes including external knowledge in the outputs. This metric gives a low score if there's any external knowledge (information outside the document) in the presentation.\\
7. \textbf{Attribution Quality}: For baselines with slide attributions pointing to parts of the document that were used to create the slide, this metric assesses the accuracy of these attributions.\\
8. \textbf{Utility}: This metric gauges how easily a user can update a generated presentation. The user can use a presentation with a high utility score without much update.

\section{Prompt for final presentation generation in GDP}\label{sec:prompt_final_slide}
Table \ref{prompt_final_slide} shows the prompt used to generate the final presentation in the GDP pipeline through the use of the LLM (see Figure \ref{fig:architecture}.
\renewcommand{\arraystretch}{1.2}
\setlength{\tabcolsep}{4pt}
\begin{table}
\centering \scriptsize
\begin{tabular}{|l|}
\hline
\makecell[l]{You are an AI assistant tasked with creating a presentation. You will be \\given some paragraphs for which you must create a slide for the\\ presentation. Following are the detailed instructions on creating the slide, \\follow them while creating the slide.\\
1. Read these paragraphs, combine them to form a slide.\\ 
2. The slide will contain a short title and bullet points.\\
3. The slide should have AT MAX 7 bullet points. Each bullet point \\should have around 15 words.\\
4. If you're given a title for the previous slide, ensure that the flow \\between the slides is maintained.\\
5. Please follow the following structure in the output.\\ 
\hspace{0.5cm} Slide Title: The slide title\\  
\hspace{0.5cm} Bullet Points:  \\
Previous Slides:\\
\#\#previous_slides\#\#
\\ \\
Text: \\
\#\#Combined_paragraphs\#\# \\
\\Slide:}\\
\hline
\end{tabular}
\caption{Prompt for final presentation generation in GDP.}
\label{prompt_final_slide}
\end{table}

\section{Prompt for G-Eval for presentation quality}
\label{sec:g_eval_prompt}
Table \ref{tab:g_eval_prompt} shows the prompt that we used for G-Eval to evaluate the presentation quality.

\renewcommand{\arraystretch}{1.2}
\setlength{\tabcolsep}{4pt}
\begin{table}
\centering \scriptsize
\begin{tabular}{|l|}
\hline
\makecell[l]{On a scale of 0-10, rate the effectiveness, clarity, and overall quality of \\the following text presentation, considering factors such as organization,\\ coherence, and the ability to convey complex ideas to the audience. \\0 is the lowest score, whereas 10 is the highest score.\\ \\ Presentation:\\\#\#presentation\#\#\\ \\Score (an integer between 0 and 10):}\\
\hline
\end{tabular}
\caption{Prompt for G-Eval to evaluate the final presentation quality.}
\label{tab:g_eval_prompt}
\end{table}

\section{Prompt for GPT-Flat}
\label{sec:gpt_flat_prompt}
Table \ref{tab:gpt_flat_prompt} shows the prompt for GPT-Flat baseline.

\renewcommand{\arraystretch}{1.2}
\setlength{\tabcolsep}{4pt}
\begin{table}
\centering \scriptsize
\begin{tabular}{|l|}
\hline
\makecell[l]{You're an AI assistant that will help create a presentation from a document. \\You will be given section heading and paragraphs in that section. Your task \\is to create a presentation with ONLY \#\#number_of_slides\#\# slides from \\the document. For every slide, output the slide title and bullet points in the \\slides. 
Please follow the following structure in the output. Do not \\output slide number. \\
Slide Title: The slide title  \\
Bullet Points:  \\
New line separated bullet points  \\ \\
Following is the document, which contains section heading and paragraphs \\under that heading. \\
----------Document Started---------- \\
\#\#document\#\# \\
----------Document Ended---------- \\ \\
Presentation (only \#\#number_of_slides\#\# slides): }\\
\hline
\end{tabular}
\caption{Prompt for GPT-Flat.}
\label{tab:gpt_flat_prompt}
\end{table}

\section{Prompt for GPT-COT}
\label{sec:gpt_cot_prompt}
Table \ref{tab:gpt_cot_prompt} shows the prompt for GPT-COT baseline.

\renewcommand{\arraystretch}{1.0}
\setlength{\tabcolsep}{4pt}
\begin{table}
\centering \scriptsize
\begin{tabular}{|l|}
\hline
\makecell[l]{You're an AI assistant that will help create a presentation from a document.\\ You will be given section heading and paragraphs in that section. Your \\task is to create a presentation with ONLY \#\#number_of_slides\#\# slides\\ from the document. For every slide, output the slide title and bullet points\\in the slides. Please follow the steps provided below. \\
1. Begin by thoroughly reading and understanding the document. Identify\\the main points, key messages, and supporting details. \\
2. Find relations between different paragraphs that could be presented in\\the same slide. \\
3. Create a high-level outline for your presentation. Identify the main\\sections or topics that you'll cover. This will serve as the skeleton for your\\slides. \\
4. Choose the most important information from the document to include in\\your presentation. Focus on key messages and supporting details that align\\with your presentation objectives. \\
5. Organize the selected content into slides, maintaining a logical flow.\\Each slide should represent a clear point or topic, and the overall structure\\should make sense to your audience. \\
6. Make sure slides are descriptive. \\
7. Presentation should have only \#\#number_of_slides\#\# slides. \\
8. Please follow the following structure. Do not output slide number.\\  
Slide Title: The slide title \\
Bullet Points: \\
New line separated bullet points \\ \\
Following is the document, which contains section heading and paragraphs\\under that heading.\\ 
----------Document Started---------- \\
\#\#document\#\# \\
----------Document Ended----------\\ \\ 
Presentation:  }\\
\hline
\end{tabular}
\caption{Prompt for GPT-COT.}
\label{tab:gpt_cot_prompt}
\end{table}

\section{Prompt for GPT-Cons}
\label{sec:gpt_cons_prompt}
Table \ref{tab:gpt_cons_prompt} shows the prompt for GPT-Cons baseline.

\renewcommand{\arraystretch}{1.2}
\setlength{\tabcolsep}{4pt}
\begin{table}
\centering \scriptsize
\begin{tabular}{|l|}
\hline
\makecell[l]{You're an AI assistant that will help create a presentation from a document. You \\will be given section heading and paragraphs in that section. Your task is to create \\a presentation with ONLY \#\#number_of_slides\#\# slides from the document. For \\every slide, output the slide title and bullet points in the slides. Please follow the\\ steps provided below. \\
1. Begin by thoroughly reading and understanding the document. Identify the \\main points, key messages, and supporting details. \\
2. Find relations between different paragraphs that could be presented in the \\same slide. \\
3. Create a high-level outline for your presentation. Identify the main sections or \\topics that you'll cover. This will serve as the skeleton for your slides. \\
4. Choose the most important information from the document to include in your \\presentation. Focus on key messages and supporting details that align with your \\presentation objectives. \\
5. Organize the selected content into slides, maintaining a logical flow. Each \\slide should represent a clear point or topic, and the overall structure should make \\sense to your audience. \\
6. Make sure slides are descriptive. \\
7. Presentation should have only \#\#number_of_slides\#\# slides. \\
8. Each slide should have around 7 bullet points. Each bullet point should have \\around 15 words.\\
9. Please follow the following structure. Do not output slide number.\\  
Slide Title: The slide title \\
Bullet Points: \\
New line separated bullet points \\ \\
Following is the document, which contains section heading and paragraphs under \\that heading.\\ 
----------Document Started---------- \\
\#\#document\#\# \\
----------Document Ended----------\\ \\ 
Presentation:  }\\
\hline
\end{tabular}
\caption{Prompt for GPT-Cons}
\label{tab:gpt_cons_prompt}
\end{table}

\section{Prompt for creating clusters in GDP-GPT}
\label{sec:gpt_decouple_prompt}
Table \ref{tab:gpt_decouple_prompt} shows the prompt for creating clusters in GDP-GPT

\renewcommand{\arraystretch}{1.2}
\setlength{\tabcolsep}{4pt}
\begin{table}
\centering \scriptsize
\begin{tabular}{|l|}
\hline
\makecell[l]{You're an AI assistant that will help create an outline for the presentation from a \\document. You will be given section headings and paragraphs in that section. Your \\task is to cluster paragraphs that should be present in the same slide. Each paragraph \\will have a unique id. The ordering of these clusters represents the order of the slides. \\Please note that there will be EXACTLY \#\#number_of_clusters\#\# clusters/slides. \\\\Please follow the steps provided below.\\
1. Start by thoroughly reading and understanding the document. Gain insights into \\the main topics, themes, and flow of information.\\
2. Identify the main ideas or topics covered in the document. These will likely serve \\as the primary themes for your slides.\\
3. Group paragraphs that are closely related or contribute to the same main idea. \\Consider the logical flow of information and how topics are connected.\\
4. Arrange the groups of related paragraphs in a logical order. This order will \\determine the sequence of your slides.\\
5. Try to include multiple paragraphs in the cluster. \\
6. Ensure non-linearity in these clusters i.e. paragraphs in a cluster need not occur \\together.\\
7. Please follow the following structure in the output:\\
    Cluster Number\\
    A comma separated list of paragraph id. (Only output paragraph numbers, \\no text)\\
7. There should be exactly \#\#number_of_clusters\#\# clusters/slides.
\\ \\Following is the document, which contains section headings and paragraphs under \\that heading.\\
----------Document Started---------- \\
\#\#document\#\#\\
----------Document Ended---------- \\\\
Clusters (exactly \#\#number_of_clusters\#\# clusters/slides):}\\
\hline
\end{tabular}
\caption{Prompt for creating clusters in GDP-GPT}
\label{tab:gpt_decouple_prompt}
\end{table}


\end{document}